\documentclass[letterpaper, 10 pt, conference]{ieeeconf} 
\usepackage[utf8]{inputenc}
\usepackage{CJKutf8}
\title{HATSUKI : An anime character like robot figure platform with anime-style expressions and imitation learning based action generation}

\author{Pin-Chu Yang$^{1}$, Mohammed Al-Sada$^{2}$, Chang-Chieh Chiu$^{3}$, Kevin Kuo$^{4}$, Tito Pradhono Tomo$^{5}$,\\ Kanata Suzuki$^{6}$, Nelson Yalta$^{7}$, Kuo-Hao Shu$^{8}$ and Tetsuya Ogata$^{9}$ 
\thanks{$^{1}$Pin-Chu Yang,$^{2}$Mohammed Al-Sada, $^{5}$Tito Pradhono Tomo, $^{6}$Kanata Suzuki, $^{7}$Nelson Yalta, and $^{9}$Tetsuya Ogata are with Waseda University, Tokyo, Japan %
	\newline
        {\tt\small $^{1}$kcy.komayang@akane.waseda.jp}\newline
        {\tt\small $^{2}$alsada@dcl.cs.waseda.ac.jp}\newline
        {\tt\small $^{5}$tito@toki.waseda.jp }\newline
        {\tt\small $^{6}$suzuki.kanata@jp.fujitsu.com}\newline
        {\tt\small $^{7}$nelson.yalta@ieee.org}\newline
        {\tt\small $^{9}$ogata@waseda.jp}
        }%
\thanks{$^{1}$Pin-Chu Yang, $^{3}$Chang-Chieh Chiu, $^{4}$Kevin Kuo, $^{7}$Nelson Yalta and $^{8}$Kuo-Hao Shu are (also) with Cutieroid Project:  www.cutieroid.com
    \newline
    	{\tt\small $^{3}$kyumasaki@gmail.com}\newline
    	{\tt\small $^{4}$freemonk6436@gmail.com}\newline
    	{\tt\small $^{8}$hudmc2000@gmail.com }
        }%
\thanks{$^{2}$Mohammed Al-Sada is also with Qatar University, Doha, Qatar}
\thanks{$^{6}$Kanata Suzuki is also with Fujitsu Laboratories LTD, Japan}
\thanks{This paper is accepted by RO-MAN 2020}
}
\usepackage{graphicx}
\usepackage{hhline}
\IEEEoverridecommandlockouts
\begin{document}
\bstctlcite{IEEEexample:BSTcontrol}
\maketitle

\begin{abstract}
%Japanese character figurine culture is very popular in worldwide, especially anime based character. Although numerous entertainment robots have been developed, less have focused on embodying the Japanese character figurine. Therefore, we take the first steps to bridge this gap by developing Hatsuki, which is a humanoid robot platform that is designed to resemble common anime characters. Hatsuki novelty lies in its ability to convey 2D facial expressions and it is designed with aesthetics based on "Mecha-musume" art-style, which are prominent features of Japanese anime character designs. We explain our technical implementation (Add some information 1 sentence), and show some applications using imitation learning. We evaluated Hatsuki through a survey that investigated initial impressions and opinions about the design. The result shows that people (which includes anime otakus, anime specialist and robotics researcher) perceive Hatsuki as a blend of figure and robot, and the participant are basically praise for the design idea. Lastly, we present our future work.

Japanese character figurines are popular and have pivot position in \textit{Otaku} culture. Although numerous robots have been developed, few have focused on otaku-culture or on embodying anime character figurine. Therefore, we take the first steps to bridge this gap by developing Hatsuki, which is a humanoid robot platform with anime based design. Hatsuki's novelty lies in aesthetic design, 2D facial expressions, and anime-style behaviors that allows it to deliver rich interaction experiences resembling anime-characters.
We explain our design implementation process of Hatsuki, followed by our evaluations. In order to explore user impressions and opinions towards Hatsuki, we conducted a questionnaire in the world's largest anime-figurine event. The results indicate that participants were generally very satisfied with Hatsuki's design, and proposed various use case scenarios and  deployment contexts for Hatsuki. The second evaluation focused on \textbf{imitation learning}, as such method can provide better interaction ability in the real world and generate rich, context-adaptive behaviors in different situations. We made Hatsuki learn 11 actions, combining voice, facial expressions and motions, through neuron network based policy model with our proposed interface. Results show our approach was successfully able to generate the actions through self-organized contexts, which shows the potential for generalizing our approach in further actions under different contexts. Lastly, we present our future research direction for Hatsuki, and provide our conclusion.
\end{abstract}

\section{Introduction}
\label{sec:intro}
The Japanese term \textit{Otaku} refers to a person who is a fan of a specific subculture,  yet such term has become synonymous with people who are fans of Japanese anime (animated cartoons), manga (comics), and video games \cite{otaku1, otaku2}. Overall, the \textit{Otaku} culture has become a worldwide phenomenon, fostering many local communities, societies and global events revolving around related hobbies \cite{otaku2},\cite{internationalanime}.

An essential aspect of the \textit{Otaku} culture is figurines, which present a physical embodiment of virtual characters from Japanese anime, manga or video games. These figurines resemble 2D-like facial features that are commonly found in Japanese anime and manga designs. The rising global popularity of the \textit{Otaku} cultures and advancements in mass-production made figurines  highly desired items by fans of the \textit{Otaku} culture worldwide. 

Despite the massive popularity of \textit{Otaku} culture worldwide, we believe robotics had a minimum contribution to such a culture. There is a scarcity of research literature that investigated potential applications of robotics within the \textit{Otaku} culture. Especially, we believe that potential applications can span beyond previously investigated applications of entertainment robots, where they can directly contribute to business profitability and value creation \cite{matsumura}, similar to figurines.

\begin{figure}[t!]
\centering
\includegraphics[height=52mm]{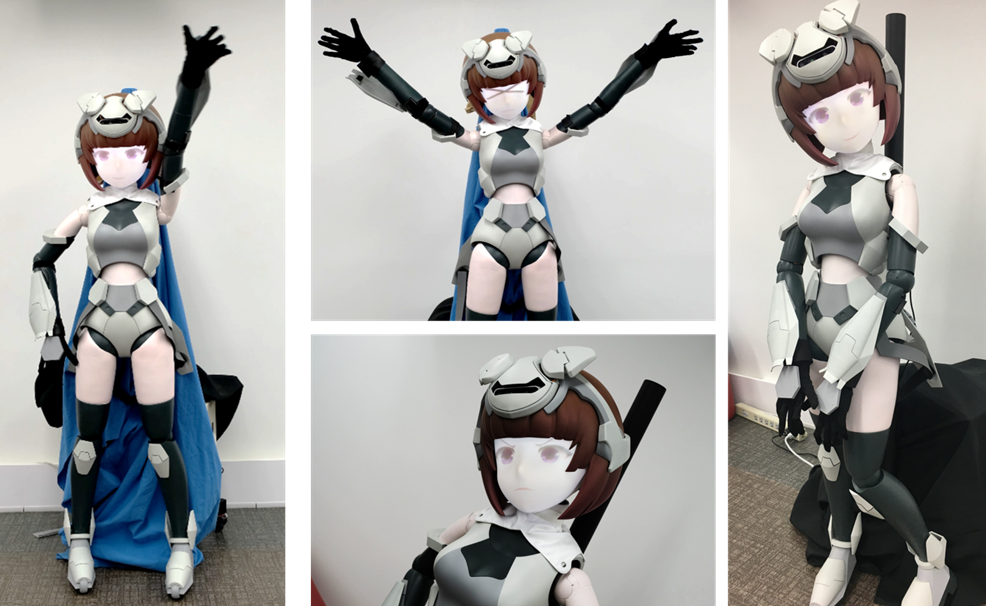}
\caption{Hatsuki is an interactive humanoid robot design that embodies Japanese anime characters, which allows for various entertainment-based interaction scenarios.}
\label{fig:hatsuki}
\vspace{-5mm}
\end{figure}

In this work, we introduce Hatsuki, which is a humanoid robot that is uniquely designed to resemble 2D anime designs found in Fig.\ref{fig:hatsuki}. Hatsuki bridges the gap between figurines and humanoid robots through its unique design and functional capabilities. As a platform, Hatsuki and its design process can be used to embody various anime-like characters in terms of aesthetics, expressions, and behavior.

Accordingly, we start by explaining the design and implementation specifications of Hatsuki, followed by two evaluations. Similar to previous approaches \cite{vatsal,surveyofcompanionrobots,alsadaorochi}, we focused our first evaluation on investigating users' impressions of Hatsuki through a survey, which was handed out to visitors of Hatsuki's exhibition booth at the largest figurine exhibition in the world (Wonder Festival \cite{wonfes}). Results show that participants regarded Hatsuki as a combination of a figurine and a humanoid robot, proposed various intriguing use cases of Hatsuki within \textit{public and private usage contexts} \cite{Pepper}, and were generally very satisfied with Hatsuki.

Interaction in the real world is hard to predict, and creating pre-defined robot actions for every world situation is impossible. Imitation learning is one approach to enable context-adaptive interactions for different situations, it is especially useful in enabling the system to perform actions through learned policy with contextual inputs (e.g. sensory information, motor information or any internal states) and  to perform various behaviors. Moreover, this approach does not require pre-defining every action-state situation, but directly learns from operators' experience to generate a policy in order to perform human-like behavior. 

Our second evaluation focuses on performing imitation learning for eleven expressive actions of Hatsuki, which were acquired through kinesthetic teaching. The results indicate that the trained neuron networks based policy successfully generates the actions and self-organize the context neurons for each different trajectory.

Lastly, we conclude that Hatsuki's evaluation results were very encouraging to pursue future works. We highlight a number of future research directions that allow Hatsuki to be applicable within a variety of novel interactive contexts of use.

We summarize the main contributions of our work as follows: 1) Design and implementation of Hatsuki, which embodies anime-character designs into an interactive humanoid robot. 2) Evaluation results that explored overall impressions of Hatsuki, and applicability of imitation learning for use in different interactive contexts.

\section{Related Works}
\label{sec:related-works}
% Think about the diagram
Our work extends three strands of related works on \textit{Humanoid Robots}, \textit{Animatronics}, and \textit{Entertainment Robots}. We discuss each of these domains as follows:

\textit{Humanoid robots} such as Twendy-one \cite{twendyone}, Asimo \cite{asimo} are designed for in-door daily life support. A subcategory of these robots attempts to resemble realistic human-designs. For example, Gemiroid \cite{gemiroid} and Sophia \cite{sophia} presented very realistic human-like appearance. Such an approach requires comprehensive design, makeup skills and integration efforts to design every aesthetic detail. Hatsuki takes a different approach as it is based on anime-character figurines. In addition, unlike mentioned works that emphasize daily life services, Hatsuki is designed to emphasize entertainment applications related to the \textit{Otaku} culture.

The design direction of Hatsuki is similar to \textit{"Animatronics"}, which are electro-mechanically animated robots that aim to mimic life-looking characters or creatures \cite{animatronics}. Various previous efforts presented vivid robots, such as humans \cite{animatronics} or animals \cite{fish} for the entertainment industry. Similarly, Hatsuki shares similarities with the works in animatronics, yet extend such works through novel aesthetic design and behaviors that mimic Japanese anime characters beyond existing works.

\textit{Entertainment Robots} is a subcategory of robots that are mainly concerned with applications like singing, dancing and various performances \cite{entertainmentRobot}. For example, Kousaka Kokona is an adult-size humanoid robot designed for entertainment, like singing and dancing \cite{Speecys}. Similarly, other robots \cite{sagawa} provide similar functionalities in smaller body proportions. Although some of the mentioned robots (e.g. \cite{Speecys,sagawa}) are designed with doll-like aesthetics, these robots are limited; they lack anime-like facial expressions, speech or autonomous interactivity. On the contrary, Hatsuki advances the state of the art by its superior design and interactive modalities, like speech, facial expressions and body gestures. Therefore, Hatsuki presents a thorough embodiment of anime-character designs beyond previous works, thereby providing various novel interaction potentials. The novelty of Hatsuki can translate to profitability and create value to consumers, in a similar fashion to previous efforts within entertainment robots \cite{matsumura}.

%Although have shown such robots can have indirectly contribute to profitability, such as by attracting more potential customers,
%a direct positive impact on the business Studies shows visited tourist increased where place introduce the robot that was original from anime works\cite{matsumura}.

%Humanoid robots and task they do (Twendy 1, Gemnoid...etc). How Hatsuki is designed a little bit differently than traditional humanoid robotsThe design of those robots are made to support daily life applications, such as ...etc (Mention some apps and references). Entertainment robots, including Hatsuki, is mainly aimed at entertainment applications with minimal life-support tasksk. Another huge difference is in terms of aesthetics, as Hatsuki is designed to mainly resemble anime figurines.

%Therefore, Hatsuki extends the basic approach in Kokona through rich anime-like facial expressions,  voice communications, and various sensors to enable various social interactions.

\section{Design and Implementation of Hatsuki}% All section Koma
\label{sec:approach}
The design approach of Hatsuki uses the \textit{outside-in} \cite{Kim2010} design approach; which refers to an aesthetics-orient design process. For usual engineering product designs, \textit{inside-out} approach is considered easier to apply due to its functional-oriented design, which starts by designing functional components of the system, followed by designing aesthetic aspects. On the contrary, the \textit{outside-in} process starts with emphasizing the aesthetic design of the robot, then proceeds to implement the technical/mechanical design of the robot in an iterative fashion. Overall, our outside-in design process is iterative and combines CAD and engineering design as well as common high-polygon 3D modeling designs. % The design flow is shown as Fig.\ref{fig:designflow}. 
Accordingly, we adopt an \textit{outside-in} design process to design and implement the various components of Hatsuki. First, we discuss the appearance of Hatsuki, followed by the facial expression system. Next, we introduce our implementation of mechanical and structural components, control and action generation in Hatsuki. 

\subsection{Appearance}
\begin{figure}[ht]
\centering
\includegraphics[height=58mm]{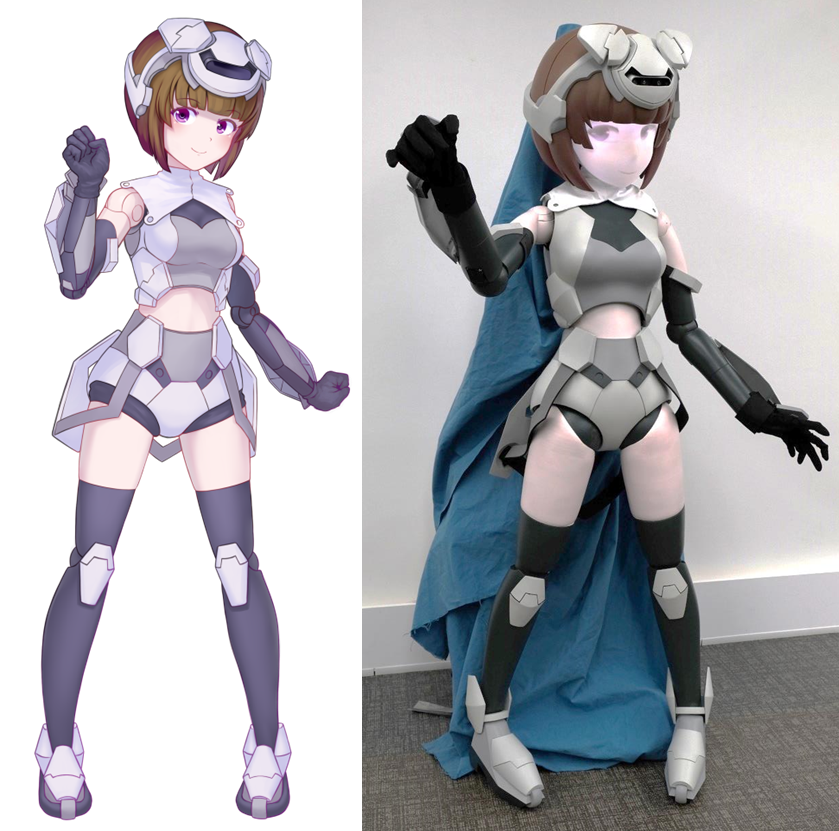}
\caption{Hatsuki extends popular anime culture character designs by embodying a \textit{Mecha-Musume} character model. Such design direction combines mechanical and anthropomorphic attributes into the aesthetics of the character. }
\label{fig:HatsukiDesign}
\end{figure}
A Japanese anime character usually applies simplified 2D characteristics, which people use to distinguish a character and largely favored in \textit{Otaku culture} \cite{May2015}. These characteristics are hairstyle, hair color, eye shape, pupils style, pupils color and eye's high-light style, especially for the main protagonist of the story who usually be designed delicately \cite{CC2012s}. Characters who do not rely on mentioned characteristics are usually recognized through unique clothes or decorations design to improve distinguishability (e.g. special hairpin on specific position). We considered mentioned aspects to design Hatsuki and make her unique and distinguishable. 
%Hatsuki  body design) We want HATSUKI mk.1 to be an animated character in appearance as a robot, but retains the cute robotic elements of animation style. so we took inspiration from the anime figures and doll parts and added some geometric designs as a basis to present a modern and a bit of industrial line without losing the original cute anime figure concept as a cool collectible item. we also want that hatsuki is a highly expandable and customized collectible item from the basic shape and appearance, so we make hatsuki's appearance more neutral, and this is also reflected in the color scheme of the body.

\begin{figure}[h!]
\vspace{-4mm}
\centering
\includegraphics[height=60mm]{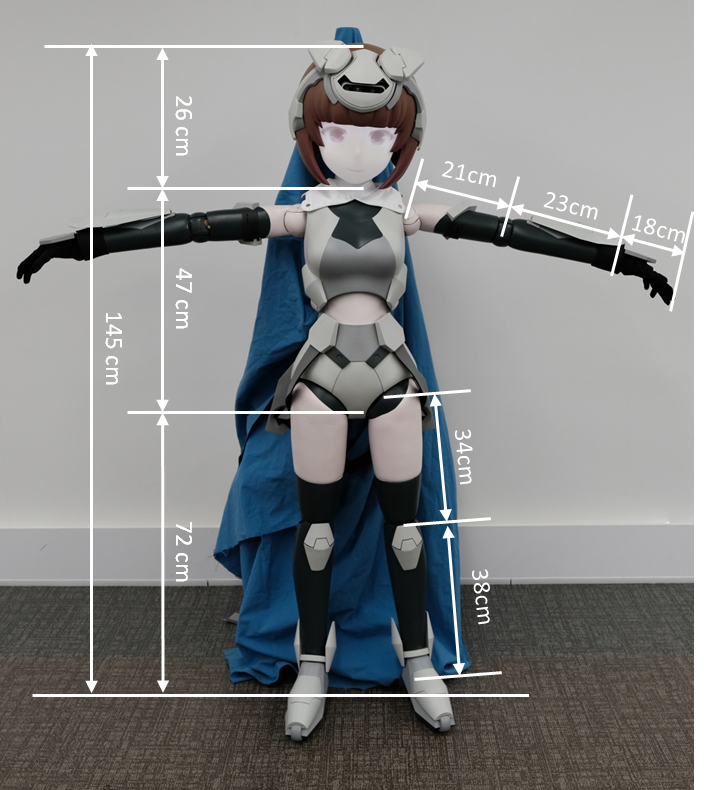}
\caption{Hatsuki has short brown hair, purple eyes, 145 centimeter tall, 1:5.5 \textit{body proportions}\cite{CC2012s2} (head-to-body ratio), which resembles a common anime-character design attributes\cite{Galbraith2009}. 
}
\label{fig:dimension}
\vspace{-4mm}
\end{figure}

The art style of design applies "Mecha-Musume", which is a popular category in \textit{Otaku culture}. This art style refers to a character that is female-like with mechanized design decoration or body parts. This art style blend mechanized design with a humanoid character, which can enable people to recognize Hatsuki as a non-human character. The final character design compared to the actual finished prototype is shown as Fig.\ref{fig:HatsukiDesign} and the appearance and dimension of Hatsuki is shown as Fig.\ref{fig:dimension}.

\subsection{Facial Expression}
\begin{figure}[h!]
\centering
\includegraphics[width=60mm]{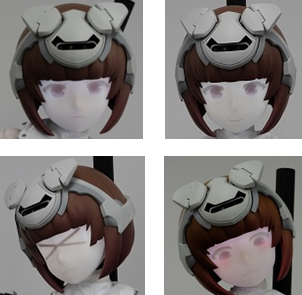}
\caption{Hatsuki use a rear projection facial expression system to provide rich anime-like facial expressions. Unlike physical facial expression mechanism (e.g. \cite{serviceHotelIshiguro}), our system is able to express a further variety of facial expressions without limitation. }
\label{fig:facial}
\end{figure}

Common anime character's facial expressions hugely vary, from human-like to abstract expressions and patterns, and such facial expressions represent a character's mind status directly \cite{CC2012s}. This characteristic is important for enriching the character's personality.  Therefore, we designed a wide variety of facial expression which are common in anime characters. A rear projection is integrated to Hatsuki to project the expressions in 2D onto Hatsuki's face as in Fig.\ref{fig:facial}. 

 % another important advantage of projection is we can change the face of the character as well, such as to customize it to other characters, add more facial features...etc
 
We apply projection mapping to project facial expression to a 3D organic face screen. Unlike \cite{Kuratate2011} trying to apply real human textures, Hatsuki's facial expression is essentially 2D animation which does not suffer from mesh un-matching issue that much.

\begin{figure}[h!]
\centering
\includegraphics[height=44mm]{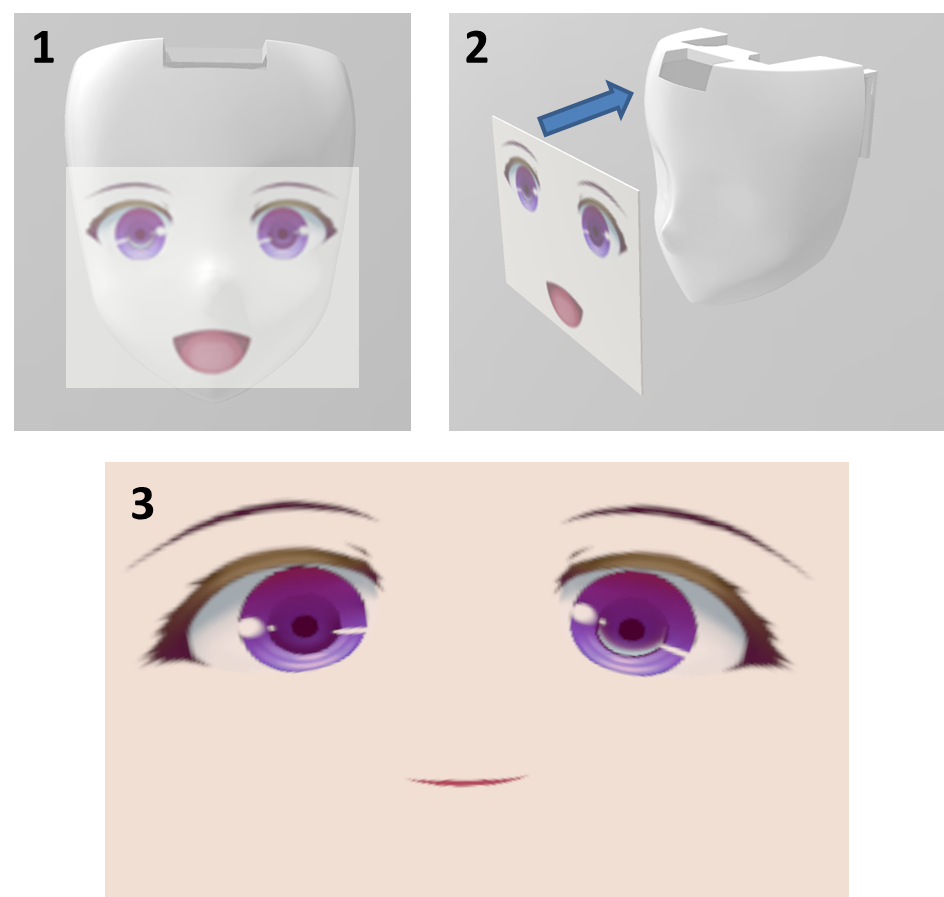}
\caption{Calibration of facial expression on 3D organic face: (1) First, we trace drawing 2D facial illustrate and animation with front view of the face 3D model. (2) After creating facial material, we do texture projection on the 3D mesh. (3) Recapture the projected texture from the projector lens position and output image to the projector.}
\label{fig:projection_mapping}
\vspace{-5mm}
\end{figure}

The calibration of projection mapping is shown as Fig.\ref{fig:projection_mapping} which applied with a parameter controllable facial texture model to perform vivid facial expressions.

\subsection{Mechanical Structure and Sensing}
Hatsuki's body is constructed using 3D printed PLA parts (Polylactic Acid). We chose PLA as it is lightweight yet robust enough to withstand the weight of various body parts. Our current implementation focuses on the upper torso design and motions. Therefore, Hatsuki currently (Mk.I version) has 17 DoFs in total, where we used a variety of servo motors to actuate different sections of Hatsuki (Table \ref{dynamixeltable})). We explain each of these sections as follows:

%The servo motors we are using is Dynamixel XM540, XM430 series motors because of their size, and they provide reasonable torque and were successfully used in imitation learning studies \cite{Ito2018}. 

% servo for arms & head

\textit{Head and Arms:} We used three servomotors (XM430) to actuate the head as well as each arm. The shoulder joint consists of servo XM540, which provides higher torque that can be used for lifting or holding objects.%(6 6 3 2)

% servo for hands % ears
\textit{Fingers:} To actuate the fingers, we use Futaba S3114 RC servo motors connected directly to the Arduino Nano's PWM pins (Pulse Width Modulation). We implemented a tendon-driven mechanism to achieve a human-size hand. Each tendon is attached to a servo motor head, where its position can be controlled by changing the PWM signal of Arduino Nano directly from a PC via serial communication. 
\textit{Ears}: We use two HobbyKing HK15148 analog servo motors (also connected to an Arduino Nano) to actuate the ears. Therefore, the ears can be used as an additional modality to communicate emotions, such as happiness or sadness.

Overall, we selected the above servo motors because they offered good trade-offs between their size and provided torque. Moreover, we chose Robotis servomotors as they provide advanced controls (such as PID), and were successfully used in imitation learning studies \cite{Ito2018}. Lastly, the Robotis servo motors communicate via RS-485 (4-wires) and can be daisy-chained through serial communication. This set up drastically enabled good overall cable management within the Hatsuki's confined design.

\textbf{Control:} The robot is controlled via a PC using an RS-485 to a USB converter (U2D2), which allows us to directly control the servomotors. We used two Arduino nano microcontrollers to control all other servomotors, which are connected to the PC using USB.% The latency time of the COM port in Microsoft Windows is 16 ms. We changed it to 1 ms so that the actuators can provide a faster response. % << did we do this?, if so how? http://emanual.robotis.com/docs/en/software/dynamixel/dynamixel_wizard2/#usb-latency-setting

\textbf{Power:} The Robotis servomotors are powered using a 12 V power supply, while the Futaba servo motors and HobbyKings servo motors are powered from a 6 V power supply.

\textbf{Sensors}: Hatsuki's head embeds an Intel RealSense D435 RGBD camera and a generic Bluetooth speaker. The Robotis servomotor also provides feedback, including position and output current sensor that enables estimating applied torque on the motors.
%%% Actuator information
%############Gripper Table#################
    \begin{table}[t]
    %\centering
    \caption{Servo Motors Specifications}\label{dynamixeltable}
    \begin{center}
    \begin{tabular}{ccccc}
    \hhline{=====}\\ [-1.5ex]
    
    Specifications              & XM430                 & XM540    & S3114  & HK15148             \\[1.0ex]
    Weight (g)                  & 82                    & 165      & 7.8    & 15            \\[0.5ex]
    Stall Torque                & 4.1(12V)              & 10.6(12V)     & 0.17(6V)    & 0.2(6V)          \\[0.5ex]
    Speed (RPM)         & 46(12V)               & 30(12V)     & 100(6V)     & 55(6V)         \\[0.5ex]
    Voltage (V)           &  10.0 - 14.8          & 10.0 - 14.8  & 4.8 - 6.0  & 4.8 - 6.0       \\[0.5ex]
    
    \hhline{=====}
    \end{tabular}
    \end{center}
    \vspace{-5mm}
    \end{table}
    %######################################
% @Koma please put the PC part: the robot is controlled by a pc, which runs Unity and basic robot software (Robotis SDK)

\subsection{Control Infrastructure}
\begin{figure}[htbp]
\centering
\includegraphics[height=50mm]{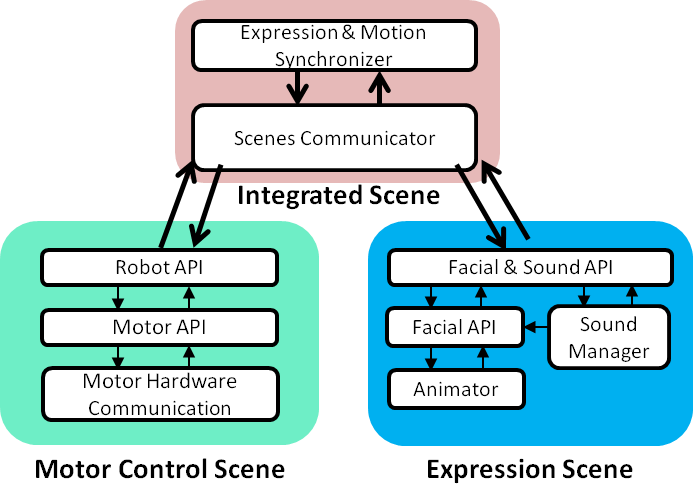}
\caption{Our control infrastructure has three main scenes, which are Motor Control Scene, Expression Scene and Integration Scene. Motor Control Scene is mainly for controlling robot actuators and related attributes. Expression Scene is for facial expression and sound system control. Integrated Scene combines the two mentioned scenes, and also provides a unified structure to develop different robot applications that combine robotic motion, voice and facial expressions.}
\label{fig:system}
\vspace{-4mm}
\end{figure}

The main controller of Hatsuki is a gaming PC, which allows us VR control or for machine learning. The control interface uses the game engine "Unity3D" due to its flexibility to develop interactive applications. 
Robotis SDK and serial communication are used to control the Robotis and Arduino nano, respectively. Our control architecture is modular, and was implemented using multiple Unity scenes as shown in Fig.\ref{fig:system}. Each scene is created to control a specific aspect of the robot, including various control parameters and attributes. New robot functionalities or sensors can easily be created or integrated by creating new scenes. Therefore, we chose a modular implementation of our system as it provides benefits in terms of development and maintainability of interactive applications.

\subsection{Action Recording}
The action refers to the performance of the robot such as motion, facial expressions, audio performance, etc. which can be observed. For motion, Hatsuki provides two style of motion recording, \textit{Kinesthetic Teaching} and \textit{record through VR device}. Facial expression and audio can be added to the recorded trajectory and create a synced command with motion. 

\subsubsection{Kinesthetic Teaching}
\textit{Kinesthetic Teaching} is a common method for teaching motor skills of the robot \cite{Akgun2012}. An operator moves the robot's body to perform motions by cutting down the torque supply for actuators and keep retrieving encoders data continuously. For most of the industrial robots, it may be hard to do kinesthetic teaching due to robot inertia, however, since Hatsuki is mostly built with 3D print material, it relatively light-weight and can be easily moved.

\subsubsection{Record through VR device}
Hatsuki also provides Virtual Reality(VR) device \textit{HTC Vive} control interface to capture the movements of an operator and transfer them to the robot. The VR trackers are used to control Hatsuki's hands, where the mapping between the VR trackers and each joint angle is first calculated through Inverse Kinematics (IK) within Unity3D, and then applied to the robot servomotors. The IK algorithm utilizes an evolutionary algorithm based method "\textit{BioIK}"\cite{Starke2017} which can create full-body, multi-objective and highly-continuous motions. 

\subsection{Implementation of Imitation Learning}
The \textit{policy} refers to a set of rules (or state) that describe how AI chooses its action to take. In this case, it represents the output model or a function that can output the action with the state input.
Imitation learning is possible to learn the policy from performing tasks \cite{Yang2017} to the dynamic motion generation \cite{deepmimic}. Such works show the advantage of learning from human operator experience capable of performing human-like behaviors while also good at interacting with an object or the environment.
By applying our neural networks based policy model, it can generate context-dependent actions that provide more variety of actions. The context can be sensory-motor information or any designed or generated contexts. 

\begin{figure}[h!]
\centering
\includegraphics[height=45mm]{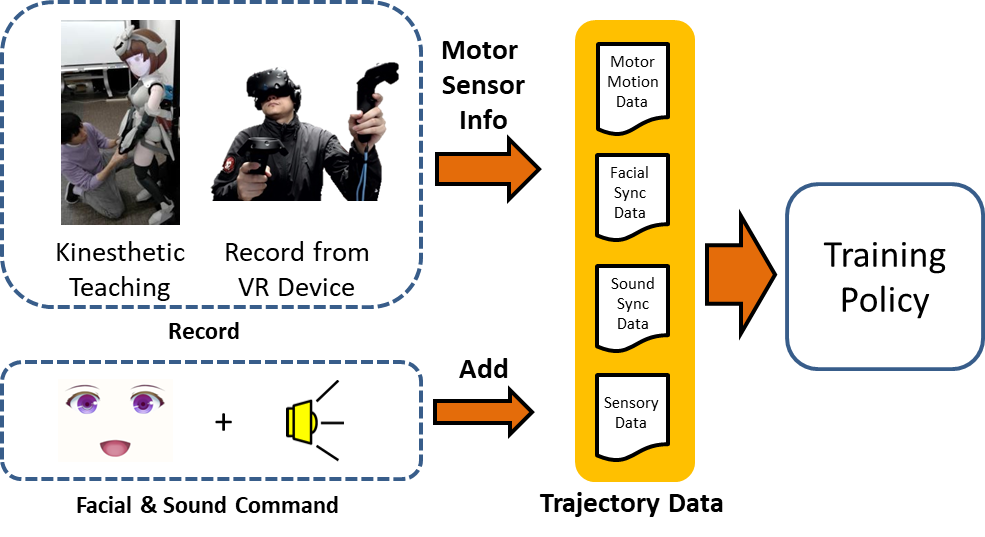}
\caption{The outline of training imitation policy }
\label{fig:TrainPolicy}
  \vspace{-5mm}
\end{figure}

The implementation of imitation learning is shown in Fig.\ref{fig:TrainPolicy} and we proceed with trained policy model embedded with the integrated scene in our control infrastructure.
For training policy, we used a multiple time-scale recurrent neural network (MTRNN)~\cite{MTRNN} due to its powerful performance \cite{Suzuki2018} to learn the relationship between facial-sound expression and motor information.
The MTRNN is composed of three types of neurons: input-output ($IO$), fast context ($Cf$), and slow context ($Cs$) neurons.The model effectively memorizes the trained sequences as combinations of the dynamics of $Cf$ and $Cs$ neurons. 

%We train a policy through trajectory with context and sensory information data use R\&P method mentioned in previous sub-section. Fig.\ref{fig:TrainPolicy} shows the outline of training an imitation policy.

\section{Evaluation 1: Investigating Impressions and Opinions toward Hatsuki}
Hatsuki's unique design and applications have various intriguing multidisciplinary challenges. First, we focus on investigating user's impressions and exception of Hatsuki. Second, we evaluate the use of imitation learning for producing different expressions.   

\subsection{Study Design and Procedure}
\textit{Objective:} To investigate the challenges and opportunities of novel robotic platforms from the user's perceptive, various previous works have utilized user-centered design approaches \cite{vatsal,alsadaorochi,Alsadachallenges,surveyofcompanionrobots} to gain insights about user impressions and expectations of such platforms. Accordingly, our main evaluation objective is to investigate the impressions and opinions about Hatsuki's design and to gain insights about highly-desired applications. We focused on how Hatsuki's aesthetics, behavior, and tasks are perceived by anime fans when compared to common anime characters and figurines. 

\begin{figure}[h!]
\centering
\includegraphics[width=70mm]{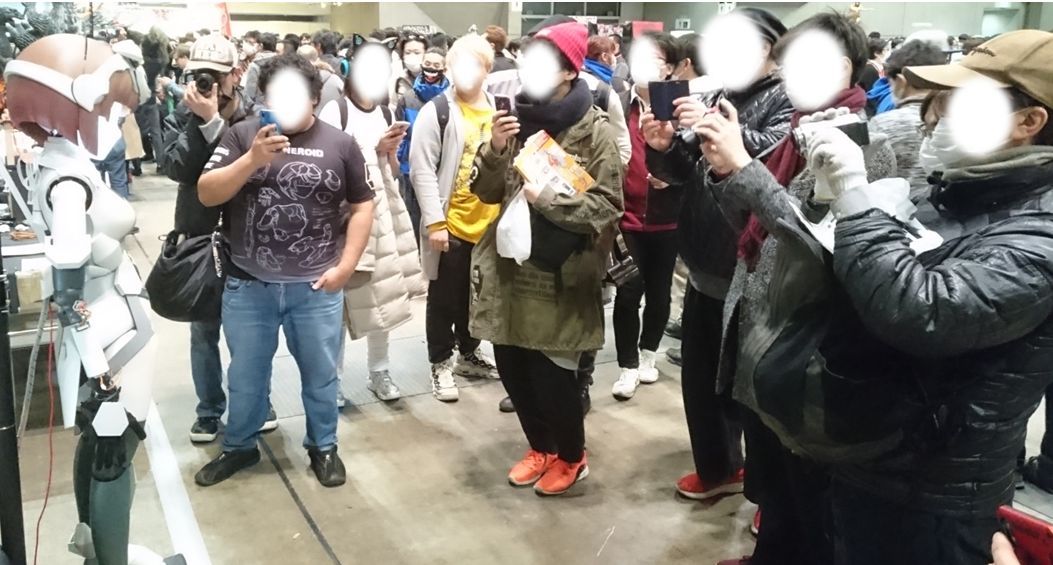}
 \vspace{-2mm}
\caption{Hatsuki demonstrate in Wonder Festival 2020 Winter event }
\label{fig:wonfesbooth}
  \vspace{-5mm}
\end{figure}

\textit{Method:}  We extended the questionnaires in previous works \cite{vatsal,alsadaorochi,surveyofcompanionrobots,QUIS} to design a survey that measured users' satisfaction of Hatsuki's design, as well as interaction expectations during public and private usage contexts. As Hatsuki is mainly designed to target anime and figurine fans, we carried out our survey at Wonder Festival 2020 \cite{wonfes}, which is the biggest anime figurine event in the world. We exhibited various interactive experiences of Hatsuki (Fig.\ref{fig:wonfesbooth}), such as greeting visitors, talking to them and posing for selfies. All experiences utilized various capabilities of Hatsuki, such as speech, facial expressions, hand gestures. 

Visitors first were given a chance to interact with Hatsuki and discuss its various capabilities with four researchers. Next, visitors were handed the survey (as described below), and researchers followed-up with visitors to ensure they understood and answered the questions correctly. Each visitor took around 10-15 minutes to complete the survey and was handed a Hatsuki seal as a reward. 

\textit{Survey:} the survey included 16 questions, that were separated into \textbf{four} sections. The \textit{first section} includes demographic questions. To ensure the participants are within our targeted audience, the \textit{second section} included questions that gauged how much each participant is into anime and figurine cultures; we asked questions about numbers of collected anime figurines and the time spent on each of mentioned hobbies. In general, participants who are fond of such culture would have many figurines and would dedicate time on a daily basis for related activities. 

Figurine culture has an associated collectability value \cite{animereport}; it is very common for people to collect and exhibit figurines of their favorite anime characters. Therefore, we wanted to know whether or not Hatsuki is perceived as a figurine, and thereby has collectability value. 

%As Hatsuki is designed based on charasteristics of common anime figurines, we asked participants whether they percieve Hatsuki as a robots or a figurine. Figurine culture has associated collectability and 
The \textit{third} section focused on understanding basic impressions of Hatsuki. Accordingly, we included brainstorming questions about the visitors' most desired use cases, whether they perceived Hatsuki as a figurine or a robot, and their overall satisfaction with Hatsuki. The \textit{fourth} section included a detailed rating of Hatsuki's various body locations. We have chosen to focus on the mentioned aspects as they provide insights about our design direction of Hatsuki and it is perceived from the viewpoint of our target audience. 

%Additional questions also investigated visitor's opinions about Hatsuki's personality while interacting (e.g. smart, cute...etc),
%four focused on initial impressions of Hatsuki, with questions gauging their we asked participants "what is the reason that made you stop at our booth",  made each visitor stop at our booth. Therefore, we asked participants about the reason of coming to our both
%Third, we asked participants "do they think Hatsuki is a robot or

\textit{Participants:} We asked 51 visitors to our booth to take our survey. Participants were aged between 20-51 (\textit{m}=41.6, females=2), Most participants were Japanese (33), while the remaining 18 came from various Asian and European countries. Participants reported spending 4.7 hours (SD=5.02) on \textit{Otaku} culture activities (anime or figurines), and reported owning an average of 31 anime figures (SD=36.32), with 30 participants owning 10 to 100 figures. Therefore, in addition to being visitors to the largest anime figurine conventions, we conclude that the participants fell into our target audience.

\subsection{Results and Analysis}
The gathered results indicate a variety of intriguing aspects regarding participants' expectations and impressions toward Hatsuki. Accordingly, we classified results based on the before mentioned survey sections and discuss them in the following subsections.

\begin{figure}[h!]
\centering
\includegraphics[width=62mm]{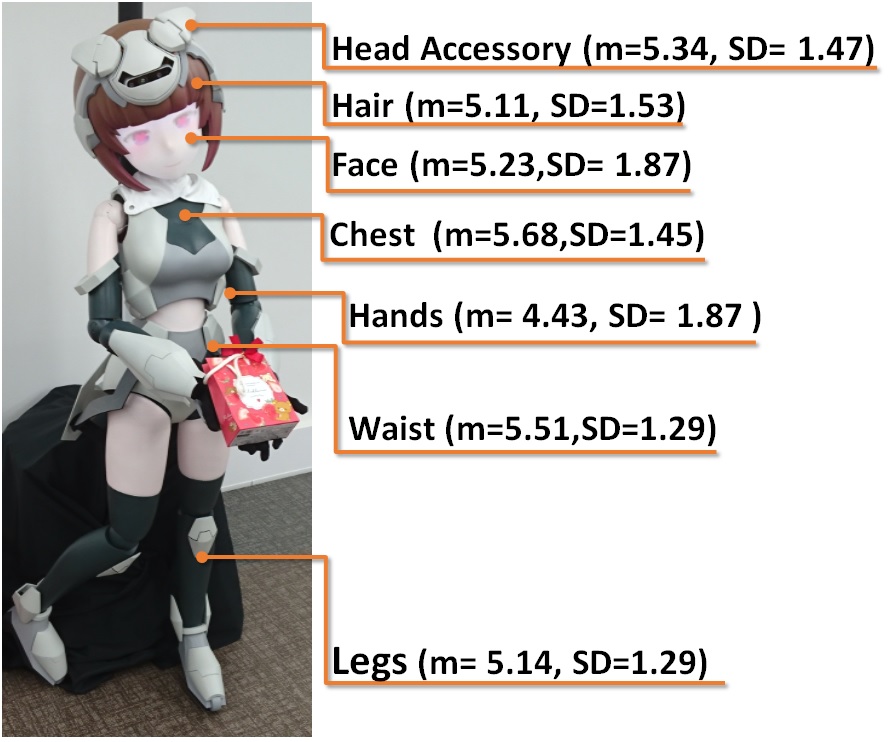}
\vspace{-3mm}
\caption{Participants' rate of Hatsuki's aesthetic design with emphasis on specific body sections (1-7 likert-scale, 7 is best) .  }
\label{fig:Ranking}
  \vspace{-5mm}
\end{figure}

\subsubsection{Impressions of Hatsuki's Design}
We asked participants to rate Hatsuki's aesthetic design (Fig.\ref{fig:Ranking}). Overall, the results indicate that participants liked Hatsuki's design (m=5.20, SD=1.55). Qualitative results reveal further insights about mentioned ratings. Participants especially praised the rich 2D expressions of Hatsuki's face, hairstyle and smooth movements. They also liked the \textit{Mecha-Musume} based anime design and body proportions, which they thought resembles cute anime figurines. 

Participants also thought some aspects of Hatsuki should be improved. Participants wanted Hatsuki to walk around the exhibition area. Also, they criticized the arms due to exposed wiring, thereby assigning a slightly lower rating. However, one-way ANOVA test with pair-wise comparisons using the bonferroni correction turned negative. Therefore, although the hand rating is slightly low, the difference is not significant. Some participants criticized the facial-expression system, mentioning it is too bright and a little pixelated. We intend to address these issues through the full implementation of a bipedal walking system in Hatsuki, better wire management and installing a better projection unit.
% above improvements can go into future work section
\subsubsection{The Use Cases of Hatsuki}
Participants detailed a variety of intriguing use cases of Hatsuki. In total, we gathered \textbf{108} use cases, and We extend classifications in previous works \cite{vatsal,alsadaorochi} with further categories to classify the use cases within the public and private contexts \cite{Pepper} as described below:

In \textit{private contexts}, such as at home, participants proposed a total of \textbf{49} use cases that we classified under four main categories. First, the majority of tasks fell under \textit{companionship applications} (\textbf{25} use cases), where Hatsuki is expected to greet users and engage in conversations about daily topics in different contexts (e.g. during dinner or before going to work). Daily life assistance (\textbf{15}) included tasks similar to common service robots, such as doing house chores (e.g. cleaning, organizing house items...etc), and providing services like waking the user up and serving food items. Participants thought Hatsuki is useful in private performances (\textbf{5}), such as singing, dancing or talk-shows. Lastly, digital assistance (\textbf{4}) comprised tasks like telling the latest news, playing music and scheduling, similar to smart home devices.

In \textit{public contexts}, such as during events or conventions,  participants proposed a total of \textbf{59} use cases that we classified under three categories. First, \textit{public performances} in events and conventions (e.g. entertainment robots \cite{entertainmentRobot}, such as dancing, singing and posing in a cute manner constituted \textbf{33} use cases. \textit{Interacting with users} in public contexts, such as by talking, hugging, shaking hands, and playing games (e.g. paper rock scissors)  included \textbf{21} use cases. Lastly, providing public services (5), included tasks like shop-keeping, receptionist and serving drinks to users. 

We asked participants about the most suitable interaction context for Hatsuki. \textbf{65.22\%} of the participants thought that Hatsuki is better suited for public contexts, citing examples of anime events, conferences and tourist attractions as potential deployment venues. Moreover, participants provided many references from pop-culture and anime characters to give examples of proposed tasks in public context, like virtual entertainment performers (e.g. Hatsune Miku \cite{hatsuneMiku}).

\subsubsection{Discussion}
%Overall, we believe Hatsuki's evaluation results are very encouraging. 
The results indicate that our design direction is highly favored, and participants provided various insights to further enhance our design direction. Likewise, the proposed use cases provided insights about potential application and deployment contexts. Participants rated their overall satisfaction with Hatsuki with \textbf{5.65} (SD=1.48, 7 is best). Therefore, we believe Hatsuki was generally well-received. A correlation analysis (Pearson product-moment correlation) to understand which aspect of Hatsuki's aesthetic ratings affected satisfaction revealed significant results for the face ({r=0.687, n=50, p\textless 0.001}). Additional tests turned negative for other body parts. Therefore, we conclude that Hatsuki's face was most significant in affecting the overall satisfaction score, which indicates the importance of designing robust facial features and expressions for this form of robots.

Although some proposed applications of Hatsuki have been investigated before (e.g. \cite{serviceHotelIshiguro,matsumura}), we believe Hatsuki advances the state of the art through its unique design; Hatsuki can be designed to embody any virtual character, from anime or pop-culture, thereby enabling experiences beyond what has been mainly investigated in social robots. For example, the familiarity of users with anime figurines characters can be used as a pretext to initiate and carry out various tasks. Such a pretext can be a significant factor in establishing familiar and trust-worthy interactive experiences.

Another interesting result is whether or not participants would consider Hatsuki a figurine with robotic components. We asked visitors to rate whether they consider Hatsuki a robot, similar to common service or companion robots, or a figurine (7 means Hatsuki is a figurine). The average response was \textbf{4.05} (SD=1.59), where most participants indicated that Hatsuki is both a figurine and a robot; since Hatsuki is designed to resemble common figurines, yet could move and interact with visitors. We believe this result verifies our design direction in Hatsuki, as it confirms that Hatsuki's design provides an appeal to visitors to pose, interact and potentially buy Hatsuki in similar to common anime figurines.

There is lack of female participants in our questionnaire, which is due to the event being mainly targeting male visitors \cite{wonfes}. Therefore, we intend to carry out a survey with other demographics, such as with predominantly female participants or in other countries. We believe such research direction would yield deeper insights about Hatsuki's appeal in varied target groups.

Overall, the results indicate that \textit{Otaku culture} fans highly appreciated Hatsuki, and provided a variety of desired tasks within public and private contexts. Therefore, we are encouraged to further use Hatsuki as a platform for research by realizing proposed tasks and deploying Hatsuki within incoming \textit{Otaku} events.

%In order to survey our target participants, We exhibited Hatsuki at Wonder Festival[ref] which is one of worlds biggest anime figurine festivals [Ref]. 
\vspace{-1mm}
\section{Evaluation 2: Imitation learning}
%\section{Evaluation 2: Evaluation through imitation learning}% Koma
We evaluated our method from the viewpoint of the imitation learning platform with a MTRNN policy. 
By using imitation learning, the proposed method enables the robot to generalize actions without designing many action details. 
This is useful for making the character more lifelike with natural motions based on observed context, rather than pre-recorded ones.
In this study, we performed imitation learning with time-series data obtained by using the Hatsuki platform. 
\par

\subsection{Experimental Setup}

\par
For training data, we obtained ten motion patterns by the kinesthetic teaching, as described in the section.III.E. The ten actions are the following: 1) self-introduction, 2) feeling challenged, 3) angry, 4)annoyed, 5)confused, 6) rejection, 7) hating something, 8) joy, 9) sad, 10) \& 11) two expressions of agreeing with the user.
Each motion pattern consists of 17-dimensional joint angles, facial expression commands, and audio commands. 
We converted facial expression and audio commands to a one-hot vector format, and incorporated them into time-series data. 
The value input to the MTRNN was scaled to [-1.0, 1.0]. 
We set parameters of our model according to the previous study~\cite{Suzuki2018}.
\par

\subsection{Results}
\begin{figure}[tbh]
  \vspace{-3mm}
 \centering
  \includegraphics[height=40mm]{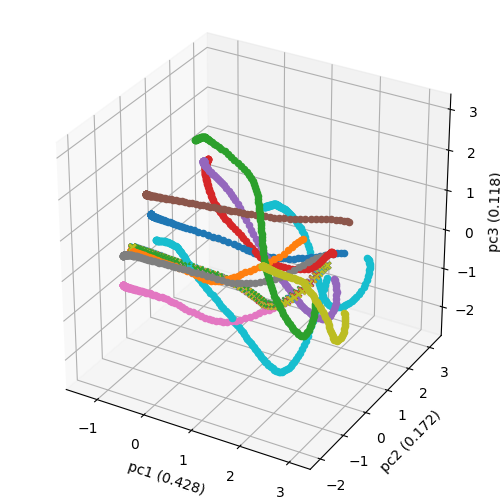}
  \vspace{-3mm}
  \caption{PCA of context neurons of ten motion patterns.}
  \label{fig: Imitation_CSPCA}
  \vspace{-5mm}
\end{figure}

Using Hatsuki, our approach successfully generated the sequence of actions, which includes motions, facial expressions and voice expressions.
The generated actions are shown as Fig.\ref{fig:genmotions} and all actions can be found in the supplementary video. 
The motion is generated by inputting the initial value of $Cs$ and the initial posture of the robot to MTRNN. 
The errors in the generated trajectories are small, indicating that the model has successfully learned high-dimensional motions. 
We also visualized the internal state of the MTRNN by principal component analysis (Fig.\ref{fig: Imitation_CSPCA}). 
Each color in Fig.\ref{fig: Imitation_CSPCA} indicates the trained sequences.
We also confirmed that each motion pattern was separated and that the MTRNN can generate robot behaviors from the initial value of $Cs$ successfully. 

\par
\begin{figure}[h!]
\centering
\includegraphics[height=40mm]{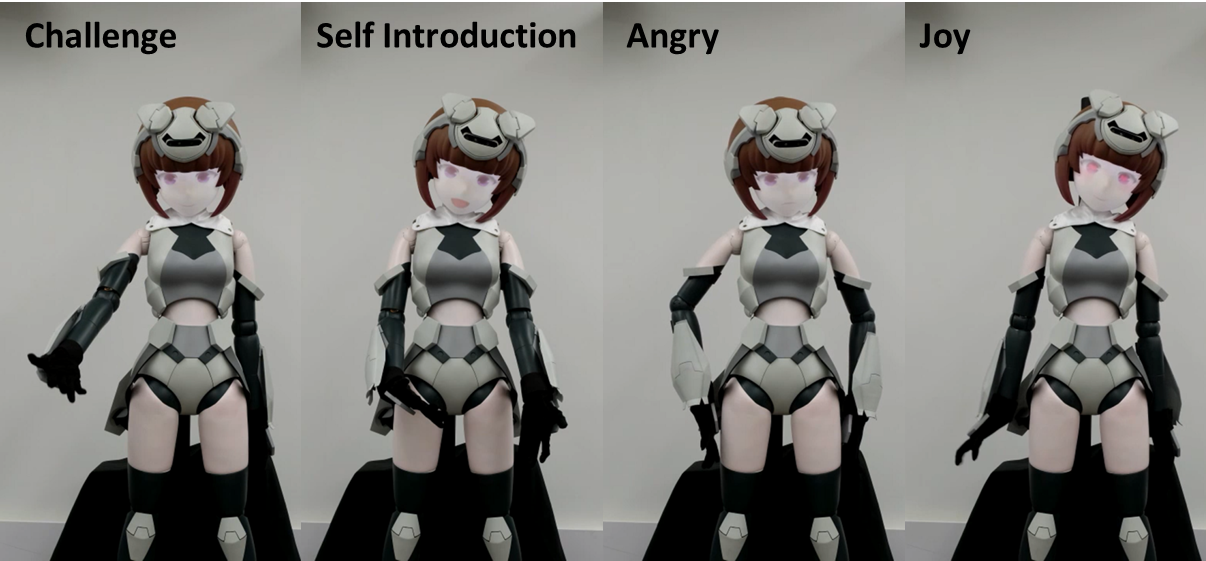}
 \vspace{-6mm}
\caption{Four samples of the motions generated through MTRNN model }
\label{fig:genmotions}
\end{figure}
%\begin{figure}[tbh]
% \centering
%  \includegraphics[height=50mm]{Figures/ImitationGen_Robot.png}
%  \caption{Online generation with Hatsuki}
%  \label{fig: ImitationGen_Robot}
%\end{figure}

We discussed the effectiveness of the proposed method from the viewpoint of the reduction in the number of work steps of imitation learning. In the conventional imitation learning methods, the process of linking the robot controller with the projection mapping, facial expression, and audio perform requires the following four processes: 
(1) Acquisition of robot motion by using the kinesthetic teaching,
(2) Synchronization of motion data, facial expression, and audio perform,
(3) Training the model of the MTRNN policy, and 
(4) Loading the trained model into the system, and synchronizing the robot motion, acquisition of sensor information, and expression of robot.

Overall, the above processes utilizes our multi-scene structure that was developed in Unity3D (Fig.\ref{fig:system}), which provides high flexibility to integrate various robot control modules. Therefore, we believe the modularity of our system reduces the system switching-cost upon deploying different control modules that is associated with conventional robotic system.

%Why we selected imitation learning + importance of imitation learning for entertainment robots
%How is our approach better than others

%Evaluation Objective: using imitation learning method, we can generalize actions, add personality to the robot using imitation actions, actroid playback. This is useful for character more life like, and its motions are not pre-recorded, but generated from observed context \textbf{[Reference]}.
%Apparatus
%Procedure: how long? why you selected these motions?
%How the actions are linked to hatsuki or entertainment applications.
\vspace{-2mm}
\section{Conclusion and Future Work}
In this paper, we presented Hatsuki, which is a humanoid robot that is designed to resemble anime figurines in terms of aesthetics, expressions and movements. We explained our implementation specifications and potential applications. We carried out an evaluation to understand user impressions regarding Hatsuki's design, as well as potential usage scenarios. We also carried out an evaluation using imitation learning which shows it can successfully perform action generation through learnt policy model effectively.

Overall, the results are very encouraging to pursue further work. We believe Hatsuki was generally liked by visitors, and they thought Hatsuki is an embodiment of anime-characters in real life. Therefore, we will focus on investigating applications and available hardware design of Hatsuki within the \textit{Otaku culture} worldwide. In the future, we intend to improve our design by implementing better robot hand for object manipulation and human interaction, as well as bipedal system so Hatsuki can walk and demonstrate full body movements.

Lastly, we believe that features like branching stories \cite{oregairu} and dating simulators \cite{loveplus} are highly sought after features of anime style games. Such features provide interactive and unpredictable elements, which were also found to be greatly liked features of entertainment robots \cite{unexpectedness}. Therefore, we intend to realize similar features in Hatsuki. Our evaluation results show the potential of imitation learning to provide adaptive behaviour to different interactive contexts, such as motions, facial expressions or speech, which can be used to provide traits similar to anime-style games. Therefore, we intend advance our work in this direction and using Hatsuki as a deployment platform.

%Moreover, since Hatsuki was designed as a \textit{Mecha-Musume} and female Otaku fans tend to be shy, we believe our first study has mainly attracted male participants to our booth. Accordingly, we intend to carry out further evaluations with different demographic groups (e.g. different countries, ethnicity or predominantly female sample).

% We will add microphone array to make Hatsuki listen and interact like speech agents...etc

%\cite{adams1995hitchhiker}
  \vspace{-2mm}
\begin{CJK}{UTF8}{min}
\bibliographystyle{IEEEtran}
\bibliography{Reference}
\end{CJK}

\end{document}